\documentclass[conference]{IEEEtran}
\usepackage{cite}
\ifCLASSINFOpdf
   \usepackage[pdftex]{graphicx}
\else
\fi

\usepackage{amsmath}
\usepackage{mathtools}
\usepackage{booktabs}
\usepackage[linesnumbered,ruled]{algorithm2e}

\usepackage{stfloats}
	
\usepackage{xcolor}

\usepackage{url}

\begin{document}
%
\title{Profiling Players with Engagement Predictions}

\IEEEoverridecommandlockouts

\author{\IEEEauthorblockN{Ana Fern\'andez del R\'io$^{1,2}$, Pei Pei Chen$^{1}$ and  \'{A}frica Peri\'a\~{n}ez$^1$}
 \IEEEauthorblockA{
 (1) Yokozuna Data, a Keywords Studio\\
  102-0074 Tokyo, Chiyoda 5F, Aoba No. 1 Bldg., Japan\\
 (2)  Deapartamento de F\'isica Fundamental,\\
 Universidad Nacional de Educaci\'on a Distancia (UNED)\\
 Paseo Senda del Rey, 9. E-28040 - Madrid, Spain\\
 \{afdelrio, ppchen and aperianez\}@yokozunadata.com}
 }

\IEEEpubid{\begin{minipage}{\textwidth}\ \\[12pt]
978-1-7281-1884-0/19/\$31.00 \copyright 2019 IEEE
\end{minipage}}

\maketitle

\begin{abstract}
  The possibility of using player engagement predictions to profile high spending video game users is explored. In particular, individual-player survival curves in terms of days after first login, game level reached and accumulated playtime are used to classify players into different groups. Lifetime value predictions for each player---generated using a deep learning method based on long short-term memory---are also included in the analysis, and the relations between all these variables are thoroughly investigated. Our results suggest this constitutes a promising approach to user profiling. 
\end{abstract}


%
\IEEEpeerreviewmaketitle

\section{Introduction}

Nowadays most video games are played online and every action by every player is recorded. This generates extremely rich datasets that---with the aid of machine learning techniques---can provide deep insights on user behavior, including accurate predictions of the future actions of each player. Increasingly diverse demographics are now playing games in a highly competitive market. Furthermore, we are clearly in the middle of a paradigm shift, as studios rely more and more in in-app purchases for income. For all these reasons, developing an in-depth knowledge of its players and their motivations is key to the success of any title. In particular, appropriately profiling players in order to meaningfully target them and cater to their needs can prove to be a game changer.

Here we propose to use lifetime, in-game progression (in terms of game level reached), playtime and purchase predictions for individual players---understood as a proxy for their expected engagement---to define meaningful segmentations (groupings of the players).

\begin{figure*}[ht!]
  \centering
  \includegraphics[width=0.33\textwidth]{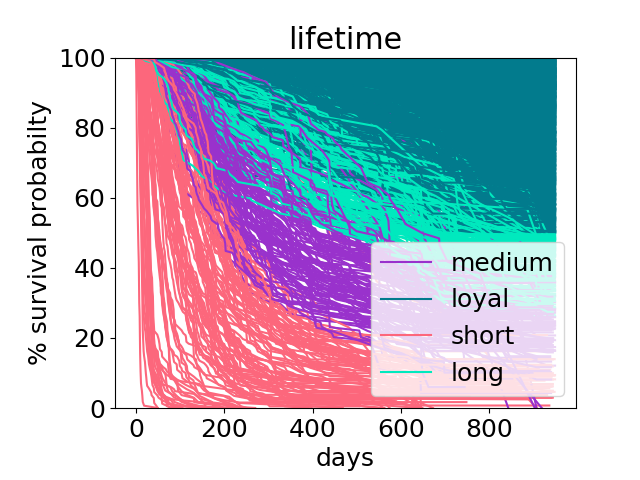}\hfill 
  \includegraphics[width=0.33\textwidth]{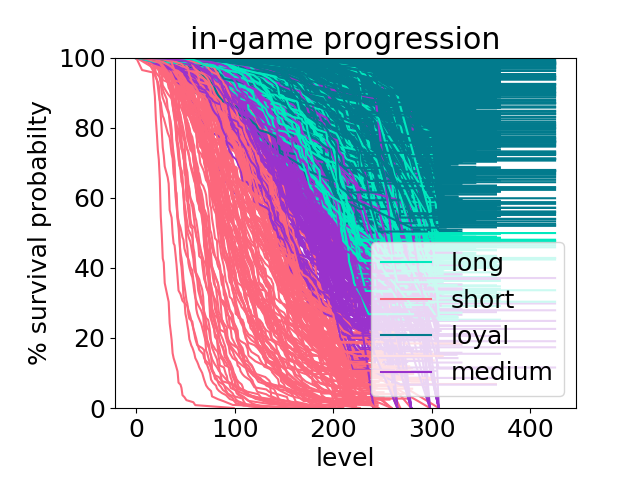}\hfill
  \includegraphics[width=0.33\textwidth]{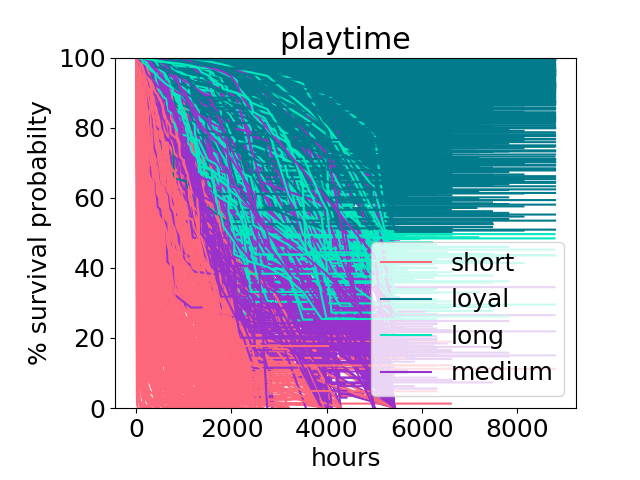}    
\caption{
Survival curves for all considered players in terms of lifetime (days since first login; left), in-game progression (game level reached; middle) and accumulated playtime (hours played; right). Colors distinguish the various lifespan groups (\emph{short, medium, long} and \emph{loyal}) for the corresponding variable.}
\label{fig:scs}
\end{figure*}

\begin{figure*}[hb!]
  \centering
  \includegraphics[width=0.25\textwidth]{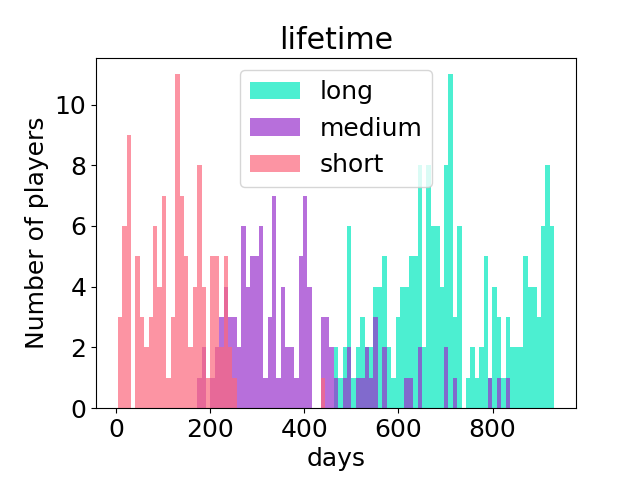}%
  \includegraphics[width=0.25\textwidth]{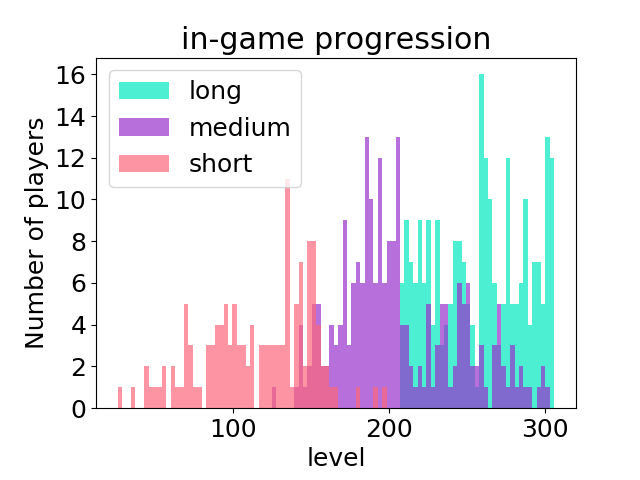}%
  \includegraphics[width=0.25\textwidth]{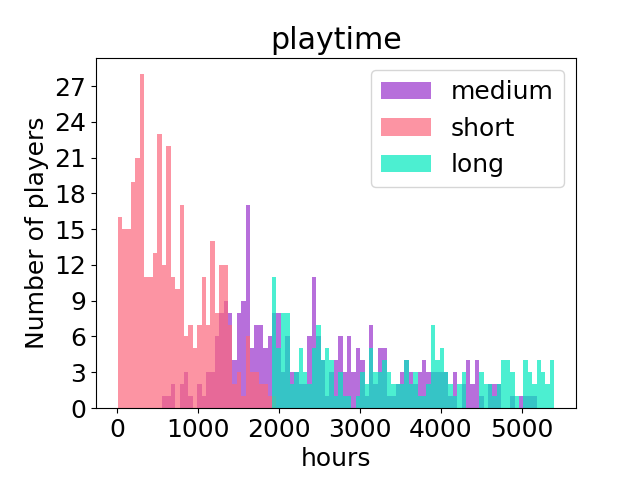}%
  \includegraphics[width=0.25\textwidth]{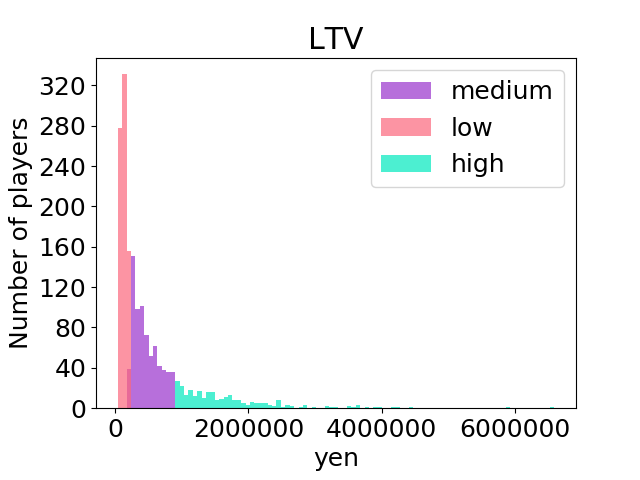} 
\caption{
Histograms of the predicted (from left to right) lifetime, in-game progression, playtime and LTV. Players are classified as described in the text, with groups shown in different colors. All players except those labeled as \emph{loyal} (for whom the median value of the survival curve does not exist) are shown.}
\label{fig:hist}
\end{figure*}

Player engagement is a complex abstract concept and a formal quantitative definition remains elusive. For the sake of segmentation, where the main goal is typically to pinpoint high-value users, outlay has been traditionally considered as the main variable---in particular in free-to-play games, where in-game purchases often constitute the main source of revenue. Player expenditure is definitely the best way to measure purchase engagement, which is of utmost importance but only tells part of story. Considering additional behavioral information on players can greatly enrich the picture and help attain a classification reflecting not only their economic value, but also markedly different playstyles, skills and interests.

To describe expected engagement we will consider the predicted values of four different variables, namely lifetime (in terms of days after first login), playtime (in hours), in-game progression (in terms of game level attained) and total expenditure or \emph{lifetime value} (LTV, in real or in-game currency). Note that all these variables are general enough to be present in virtually all titles, which makes the method proposed in this work easily generalizable to games of any kind.\looseness=-1

And not only are these variables interesting for classification purposes in themselves, but their combination also provides a deeper understanding of the different types of players in a game. For instance, players with a very rapid in-game progression (who reach a high level after a relatively short playtime, regardless of their lifetime) and low spend might be overlooked by traditional segmentation methods due to their lack of direct economic value; however, these are the most skillful players, and a careful study of their traits and behavior---allowed by our approach---could provide developers with a lot of useful insights. 

The use of predictions allow us to appropriately profile players from their very first steps in the game and to consider all available information, including that of players who are no longer in the game---but who can nonetheless help to better understand 
the active players and what can be expected of them.\looseness=-1 

Our method characterizes each player by means of a collection of survival curves that present the (past and future) probability of remaining in the game after a certain lifetime, playtime or number of levels. These three variables---which we take as indicators of player engagement---are modeled using survival ensembles, as thoroughly described in \cite{perianez2016churn,GameBigData}. Moreover, we also include lifetime value (LTV, the total spend of a player throughout their lifetime in the game) predictions, generated through a deep learning method based on \emph{long short-term memory} (LSTM).
To the best of our knowledge, this is the first work that proposes performing player segmentation using not only historic data, but also predictions of future behavior. 


\section{Models used}
\label{sec:models}

\subsection{Conditional inference survival ensembles}
Conditional inference survival ensembles \cite{hothorn2006unbiased} are ensemble learning models with survival \cite{clark} trees as the underlying algorithm. A detailed explanation of this technique, exact settings and performance can be found in \cite{perianez2016churn, GameBigData}.\looseness=-1

\subsection{Long short-term memory models}
Long short-term memory \cite{hochreiter1997long} is a recurrent neural network architecture designed to model temporal sequences and their long-range dependencies. The specific implementation used here consists of two time series---player actions and purchase logs---that are separately introduced in the model using two LSTM layers, whose outputs are concatenated and inputted to a fully connected layer followed by other three fully connected layers. The output of the last layer represents the desired LTV.

\section{Methodology}
\label{sec:method}

Therefore, after running the models, we have each player characterized by three survival curves (days after first login, game level and playtime) and a single value (LTV prediction). 

For the three variables modeled through survival ensembles, we profile users by assigning them to one of four groups based on the quartile values of their survival curves and on the relation of these to the average values for the population studied. These groups correspond to \emph{short}, \emph{medium} and \emph{long} lifespans (in any of the variables) and to \emph{loyal} players, namely users who are not in risk of engagement loss according to the model and the available data. Meanwhile, LTV predictions are used to classify players into \emph{high}, \emph{normal} and \emph{low} spending groups. This grouping can be performed in many different ways, which are more or less useful depending on the particular game and the aim of the profiling. Below we describe a simple approach that provides reasonable results (see Section~\ref{sec:results}) and is general enough to be applied to a broad range of games.  

For lifetime, playtime and in-game progression, players whose survival probability always remains above 50\% (i.e., for whom the median value of the survival curve does not exist) are labeled as \emph{loyal}. Players are considered to have \emph{long} lifespans if their ``final probability'' is greater than or equal to 25\% and smaller than 50\% (by this we mean that the survival probability eventually falls below 50\% but never drops below 25\%) and their median value is greater than or equal to the population's average. The \emph{medium} lifespan group is comprised by players (i) with final probability between 25\% and 50\% but median value smaller than the population's average; (ii) with final probability below 25\% (but not zero), provided that the value at which the 25\% survival probability is reached is greater than or equal to the population's average; and (iii) with vanishing final probability, provided that their median value is greater than or equal to the average (computed considering just players whose survival probability never reaches zero). The rest are \emph{short}-lived players. 

In regard to LTV, all players are considered \emph{medium} spenders except those with expected outlay less than half of the average expected outlay, who are classified as \emph{low} spenders, and those with at least twice the expected average, who are deemed \emph{high} spenders.

\section{Results}
\label{sec:results}

The dataset used in the present work comes from the role-playing, freemium, social game \emph{Age of Ishtaria}, developed by Silicon Studio. Only top spenders (players with accumulated outlay above a certain threshold, computed from the first two months of data 
so that around 50\% of all purchases come from players above it) are considered. The data used to train the models covers October 2014 through April 2017 and contains 3265 users. Predictions are then run for the 1771 players who had not churned yet as of May 1, 2017 (by definition, those who had connected at least once in the previous 9 days).

Figure~\ref{fig:scs} displays output survival curves for all players in terms of lifetime, in-game progression and playtime---with colors distinguishing the various groups discussed above---while Figure~\ref{fig:hist} shows histograms for the corresponding median values (when these exist, i.e., for \emph{non-loyal} players) and also for the expected LTV. We can see that the predicted values are distributed differently for each variable: the distribution is more uniform for lifetime, skewed to larger and smaller values for level and playtime, respectively, and almost monotonously decreasing for LTV.

\begin{figure}
  \centering
   \includegraphics[width= 0.45\textwidth]{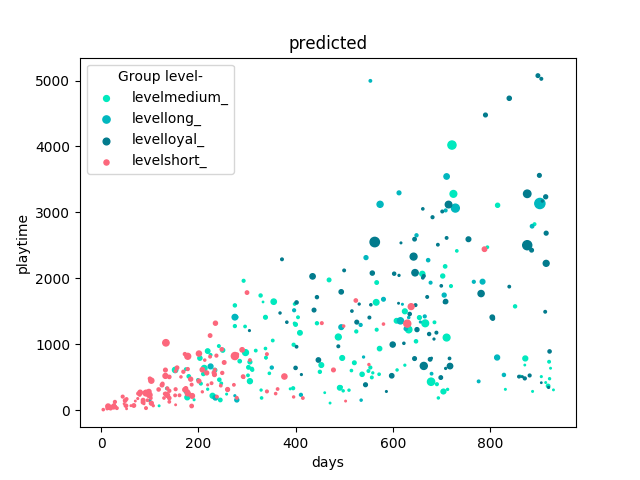} 
\caption{
Playtime versus lifetime predictions (median survival values) for all players \emph{non-loyal} in both variables. Color represents grouping in terms of predicted game level. The area of the circles is proportional to the expected LTV.\looseness=-1}
\label{fig:days-playtime-ltv-levelg}
\end{figure}

Exploring the relations among the different predicted variables can provide insights into the dynamics of the game. For instance, Figure~\ref{fig:days-playtime-ltv-levelg} compares predicted playtime (in hours) and lifetime (in days) for all players for whom median values exist, i.e., \emph{non-loyal} in both variables. As expected, the spread in playtime increases with lifetime, since periods of inactivity and differences in session lengths among players gain importance as time passes.
The area of each circle is proportional to the predicted LTV and colors represent lifespan groups in terms of game level. Although large LTVs tend to be associated to players with longer lifetimes, there are also some users with low playtime and/or lifetime and relatively large LTV, and players with small predicted outlay can be found across all scales. In regard to the expected in-game progression, there is an obvious correlation with playtime, but with variations reflecting different player skills, as shown e.g.\ by the occurrence of short level lifespans for very large playtime values (corresponding to not so skillful players). Note this kind of analysis can also point very clearly to players with extreme behaviors. For example, in Figure~\ref{fig:days-playtime-ltv-levelg} there is a player with very high expected playtime as compared to their predicted lifetime (nearly 5000 hours of playtime in less than 600 days).

\begin{figure}
\centering%
\includegraphics[width= 0.45\textwidth]{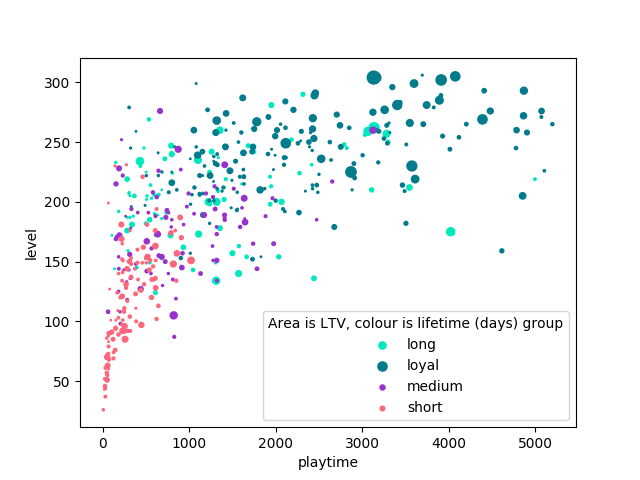}%
\caption{
Game level versus playtime predictions (median survival values) for all players non \emph{loyal} in both variables. Color represents grouping in terms of predicted lifetime. The area of the circles is proportional to the expected LTV.}
\label{fig:playtime-level-days-ltv-groups}
\end{figure}

Figure~\ref{fig:playtime-level-days-ltv-groups} further explores the playtime--level relation. Here the area of the circles is again proportional to the predicted LTV and colors represent groups in terms of expected lifetime. Progression through the first levels is clearly quick and easy (as reflected by the steep slope and relatively small spread), while for higher levels the relation between both predictions flattens and the spread becomes larger. This shows it takes longer to go through those levels and highlights the significant impact of player skills on the expected in-game progression.\looseness=-1 

\begin{figure}
  \centering
  \includegraphics[width= 0.45\textwidth]{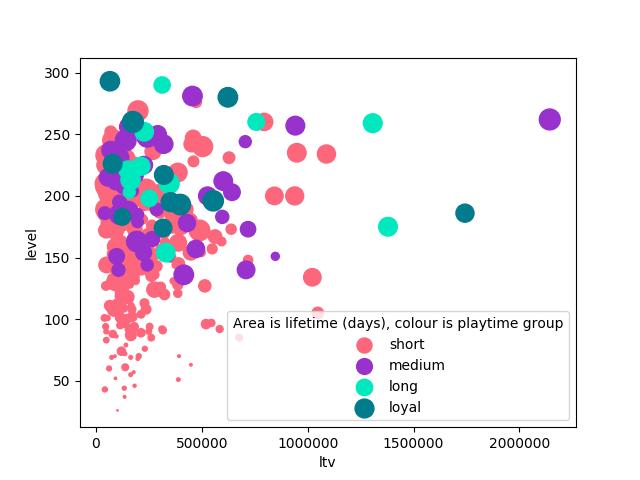} 
\caption{
Game level versus LTV predictions for all players \emph{non-loyal} in level. Color represents grouping in terms of predicted playtime. The area of the circles is proportional to the expected lifetime.}
\label{fig:ltv-level-days-playtime-group}
\end{figure}

\begin{figure*}[t!]
  \centering
  \includegraphics[width=0.33\textwidth]{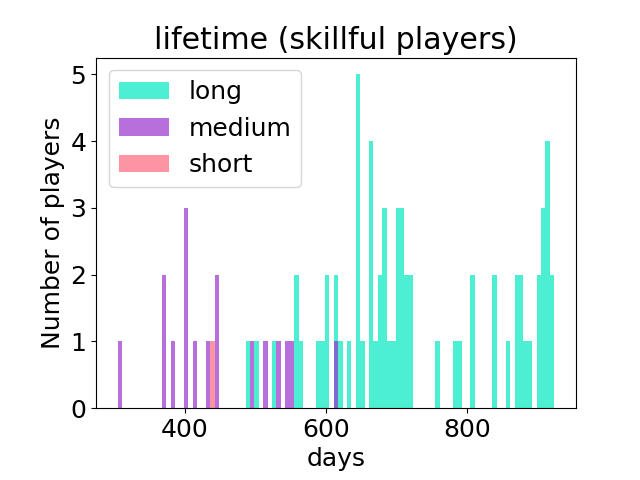}\hfill
   \includegraphics[width=0.33\textwidth]{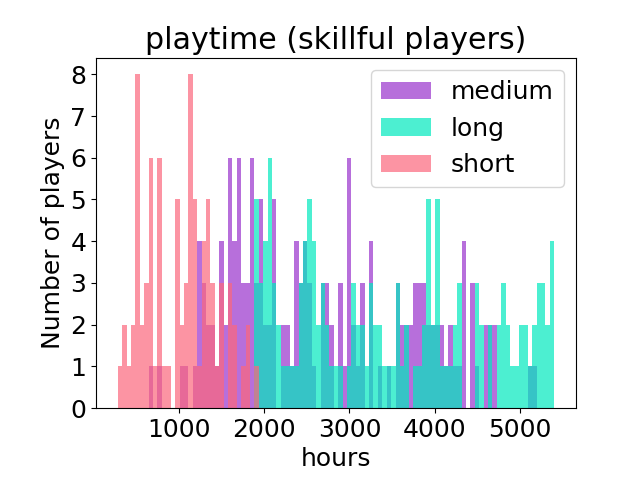}\hfill
  \includegraphics[width=0.33\textwidth]{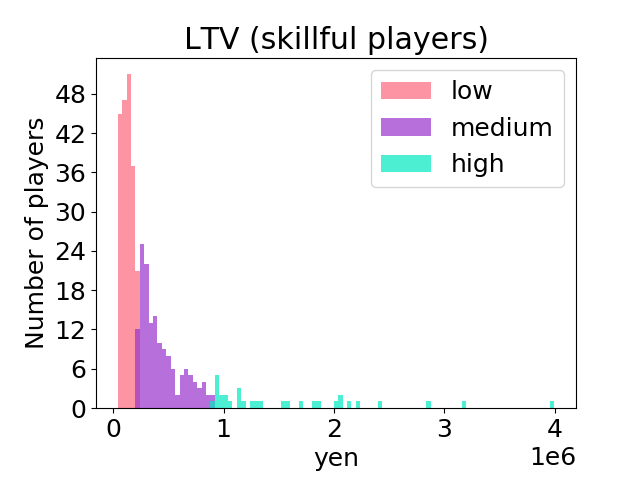} 
\caption{
Histograms of predicted lifetime (left), playtime (middle) and LTV (right) for players \emph{loyal} with respect to level and \emph{non-loyal} in terms of playtime. Colors represent different groups for the corresponding variable.}
\label{fig:dist_skilled}
\end{figure*}

\begin{figure}
  \centering
  \includegraphics[width= 0.45\textwidth]{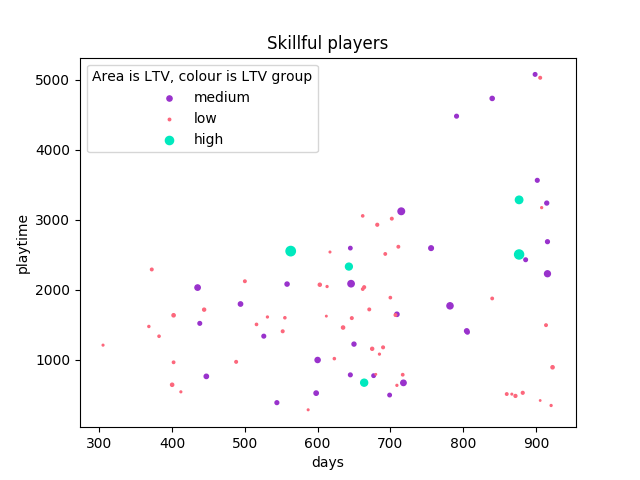} 
\caption{
Playtime versus lifetime (in days) predicted values for all players \emph{non-loyal} in both variables and \emph{loyal} in terms of level. Color represents grouping in terms of expected LTV and the area of the circles is also proportional to LTV.\looseness=-1}
\label{fig:days-playtime-ltv-ltvg-skilled}
\end{figure}

The relation between the highest level to be reached and the amount of money to be spent is studied in more detail in Figure~\ref{fig:ltv-level-days-playtime-group}, which shows that, while all players with high expected LTV will make significant in-game progression, players with small predicted outlays can do just as good. As we have seen, the correlation with playtime is not perfect either. This could point to the relative fairness of the game: while spending money \emph{might} help with in-game progression (note that the correlation could simply be showing that players who advance further in the game are likely to spend more), the latter has much more to do with the combination of skill and time spent playing.

Profiling users in terms of these variables also allow us to focus on interesting groups of players. For example, players predicted to be \emph{loyal} in terms of in-game progression (whose survival probability remains over 50\% even at high levels) and \emph{non-loyal} in playtime are conceivably the most skillful ones. The histograms of the expected lifetime, playtime and LTV for the 385 players fulfilling these conditions are displayed in Figure~\ref{fig:dist_skilled}. These variables (except for the LTV) show noticeably different patterns for this subset of skillful players as compared to the total population (cf.\ Figure~\ref{fig:hist}): now playtime is more evenly distributed and lifetime skewed to larger values. While it is normal to expect these players to have long lifetimes, interestingly, playtime predictions suggest a number of them will play frequently but for a short time or with many inactive days between sessions. Figure~\ref{fig:days-playtime-ltv-ltvg-skilled} explores the predicted lifetime vs.\ playtime relation for this group of skillful players, with the area and color of the circles related to the expected LTV. It confirms not only the pattern just described, but also the previous discussion about the relative fairness of the game: the distribution of predicted outlays does not vary significantly with respect to the global population and there are many low spending players, spread across all lifetime and playtime values.\looseness=-1

\section{Summary and Conclusion}
\label{sec:conc}

We used individual player predictions of lifetime, game level and playtime (obtained through a conditional inference survival ensembles model) and also LTV (through a LSTM-based deep learning method) to explore a simple user profiling approach, classifying players into broad categories for each of these variables. Moreover, exploring the relations between them has allowed us to analyze behavioral patterns, pinpoint outlier behaviors, draw qualitative conclusions about the game dynamics (e.g.\ about its \emph{fairness}) or focus on certain groups of players of particular interest. This suggests the proposed method constitutes a promising approach to a richer profiling landscape. Future work includes exploring different segmentation strategies, such us focusing only in detecting outliers in any/all of the variables considered, or using unsupervised time series clustering to group survival curves. 
\section*{Acknowledgements}

We thank Javier Grande for his careful review of the paper.\looseness=-1

\bibliographystyle{abbrv}
\bibliography{main.bib}	

\end{document}